\documentclass[conference,letterpaper]{IEEEtran}
\IEEEoverridecommandlockouts

\usepackage{cite}
\usepackage{amsmath,amssymb,amsfonts}
\usepackage{algorithm}
\usepackage{algorithmic}
\usepackage{graphicx}
\usepackage{textcomp}
\usepackage{xcolor}
\usepackage{booktabs}
\usepackage{subfig}
\usepackage{makecell}
\usepackage{bbm}
\usepackage{comment}
\usepackage{multirow}
\usepackage{hyperref}
\hypersetup{hidelinks}

\newtheorem{theorem}{Theorem}

\newcommand{\RAOCSRevisionColor}{black}
\newcommand{\RAOCSchanged}[1]{{\color{\RAOCSRevisionColor}#1}}

\begin{document}

\title{Joint Client Scheduling and Resource Allocation for Optimal Federated Learning over Wireless Networks
\thanks{
(1) This work was partially supported by ANR-22-CE23-0024.
(2) Some experimental simulations used GENCI-IDRIS HPC resources under Grant 2026-103464.
\RAOCSchanged{(3) An earlier version of this paper was presented at IEEE FLTA 2026.}
\RAOCSchanged{(4) The source code and instructions for reproducing the experimental results are publicly available on \href{https://github.com/AhmadDabaja/RAOCS-FLTA2026}{GitHub}.}
}
}


\author{
\IEEEauthorblockN{Ahmad Dabaja and Rachid El-Azouzi}
\IEEEauthorblockA{
CERI/LIA, Avignon Universit\'e, Avignon, France\\
Email: \{Ahmad.Dabaja, Rachid.Elazouzi\}@univ-avignon.fr
}
}

\maketitle

\begin{abstract}
    Federated learning often relies on partial client participation because full participation is costly in cross-device wireless systems with limited uplink resources, heterogeneous channel conditions, energy constraints, and strict upload deadlines. However, partial participation introduces sampling variance, causing the sampled update to deviate from the full-participation update. Existing optimal client sampling methods reduce this variance but usually rely on simplified client-count budgets, while wireless-aware scheduling methods account for communication constraints but do not directly optimize sampling variance. This paper proposes resource-aware optimal client sampling (RAOCS), a unified variance-minimizing client scheduling framework under uplink resource-block, round deadline, update-information, and energy constraints. The resulting problem admits a capped water-filling solution that balances client update importance, wireless resource cost, and energy feasibility. Experimental results demonstrate that RAOCS achieves accuracy comparable to that of OCS while reducing the average upload time per iteration by more than a factor of 252.
\end{abstract}


\section{Introduction}

Federated learning (FL) enables distributed clients to collaboratively train a global model without sharing their raw data \cite{mcmahan2017communication,kairouz2021advances}. In cross-device FL, clients often have limited computation, energy, availability, and communication resources. Therefore, practical FL systems usually rely on partial participation, where only a subset of clients is selected at each communication round \cite{mcmahan2017communication,bonawitz2019towards}.

Partial participation introduces a statistical cost. When only a subset of clients participates, the aggregated update may deviate from the update that would have been obtained under full participation, especially under non-IID data where client updates can differ significantly. To reduce this discrepancy, several works study optimal or importance-based client sampling, where participation probabilities are chosen to reduce the variance of the sampled update \cite{rizk2021optimal,rizk2022federated,chen2022optimal,wang2023delta,zhao2025adaptive}.

A representative example is Optimal Client Sampling (OCS), which minimizes the distance between the sampled update and the full-participation update under a constraint on the expected number of selected clients \cite{chen2022optimal}. Related works consider two-level agent/data sampling \cite{rizk2021optimal,rizk2022federated}, adaptive online sampling \cite{zhao2025adaptive}, diversity-aware sampling \cite{wang2023delta}, and multi-model FL \cite{zhang2024optimalvariance,zhang2025towards}. These methods provide strong foundations for variance-aware client sampling, but their communication model is typically abstracted as a client-count or sampling budget. In parallel, system-aware and wireless-aware client selection methods account for latency, bandwidth allocation, channel conditions, energy, computation time, or deadlines \cite{nishio2019client,xu2020client,huang2020efficiency,luo2022tackling,luo2024adaptive,vardhan2025client}. However, their main objective is usually to reduce wall-clock time or select feasible clients, rather than directly minimizing sampling variance with respect to the full-participation update.

This paper addresses the gap between these two directions by proposing resource-aware optimal client sampling (RAOCS) for wireless federated learning. We consider a wireless FL system with a limited number of uplink resource blocks, heterogeneous channel-dependent transmission rates, and per-round upload deadlines. Since each selected client must upload its local update before the deadline, its sampling decision depends not only on its statistical importance to the learning process, but also on the number of resource blocks required to support its transmission. The proposed framework further accounts for update-information availability through a refresh probability and incorporates client energy constraints through energy-induced sampling caps.

The main contributions are summarized as follows: \textbf{(1)} we formulate RAOCS as a variance-minimization problem under an expected uplink resource-block budget, upload-deadline constraints, update-information availability, and client-specific energy caps; \textbf{(2)} we model each client's wireless sampling cost through its channel-dependent uplink rate, computation time, and minimum resource-block requirement for deadline-feasible participation; \textbf{(3)} we derive a capped water-filling sampling rule that balances update importance, wireless resource cost, and energy feasibility, and provide a strict-budget greedy counterpart for settings requiring per-round resource feasibility; and \textbf{(4)} experiments on non-IID Fashion-MNIST, CIFAR-10, and CIFAR-100 show that RAOCS preserves OCS-level accuracy while greatly reducing upload time, with energy caps reducing energy consumption by 2.1-18.8\%.


\section{Related Work}

\subsection{Variance Optimization under Partial Client Participation}
Client selection in federated learning has been widely studied to reduce communication and computation costs under partial or intermittent participation. Many methods formulate client selection as a probabilistic sampling problem, where participation probabilities depend on the estimated contribution of each client to the global objective, often through loss reduction, update importance, or diversity \cite{rizk2021optimal,chen2022optimal,wang2023delta}.

Importance sampling has been applied to FL for both client and local-data selection \cite{rizk2021optimal}. Rizk et al. \cite{rizk2022federated} extended this to sampling without replacement using normalized inclusion probabilities, which is more representative but analytically more complex. Chen et al. \cite{chen2022optimal} introduced Optimal Client Sampling (OCS), proving that the variance-minimizing sampling probability is proportional to the client update norm. While OCS effectively reduces gradient-estimation variance relative to full participation, it is constrained solely by a client-count budget and fails to account for wireless transmission rates, resource allocation, latency, or upload-deadline feasibility. Other variance-aware methods incorporate availability \cite{ribero2022f3ast}, diversity \cite{wang2023delta}, multi-model optimization \cite{zhang2024optimalvariance,zhang2025towards}, or online bandit learning \cite{zhao2025adaptive}. However, all treat communication constraints abstractly as fixed budgets or client counts, lacking explicit wireless resource-block or upload-deadline constraints.

\subsection{Wireless-Constrained Client Selection}

A second line of work studies client selection from a system and wireless perspective. FedCS selects clients expected to complete local training and communication within a deadline, accounting for heterogeneous computation and wireless resources \cite{nishio2019client}. 
 Wireless FL methods have also optimized convergence time, bandwidth use, energy efficiency, and fairness through client selection and resource allocation \cite{chen2021convergence,xu2020client,huang2020efficiency}. Luo et al. study adaptive client sampling under system and statistical heterogeneity, including wireless latency and bandwidth, and minimize time-to-accuracy using convergence-time bounds \cite{luo2022tackling,luo2024adaptive}. Recent work also considers data heterogeneity and network latency in client selection \cite{vardhan2025client}. These approaches explicitly model wireless or system constraints, but their main objectives are system-level metrics rather than directly minimizing the variance of the sampled update under wireless resource constraints.

Several works take a communication-efficiency perspective on client selection. Ribero and Vikalo select informative updates and estimate missing ones to reduce overhead \cite{ribero2020communication}, while communication-loss trade-off methods balance upload costs against model loss \cite{hosseinzadeh2022communication}. SecOMPFL identifies high-contribution clients via compressed sensing-based global update reconstruction under a sparsity assumption \cite{li2025optimal}, but this introduces selection bias by prioritizing reconstruction accuracy over variance minimization or global-loss representation. In contrast, RAOCS directly minimizes sampled-update variance while explicitly incorporating wireless resource-block, deadline, update-information, and energy constraints.

\section{System Model and Problem Formulation}
We consider a wireless federated learning system with partial client participation. We first introduce the federated learning model and the variance induced by probabilistic sampling. We then model the update information available at the server, the uplink resource-block constraints, and the client energy constraints. These components are finally combined into a unified resource-constrained optimal client sampling problem.

\subsection{Federated Learning Model}
We consider a federated learning system with $K$ clients indexed by $k \in \mathcal{K} = \{1,\ldots,K\}$. Client $k$ owns a local dataset $\mathcal{D}_k$ of size $n_k$. The global objective is
\begin{equation}
    \min_{w} F(w)
    =
    \sum_{k=1}^{K} \alpha_k F_k(w),
\end{equation}
where $F_k(w)$ is the local empirical risk of client $k$, and $\alpha_k = n_k/n$ with $n=\sum_{j=1}^{K} n_j$.

At communication round $t$, the server broadcasts the current global model $w^t$. Client $k$ computes a local update $U_k^t$, which may correspond to one or several local SGD steps. Under full participation, the ideal global update is
\begin{equation}
    U_{\mathrm{full}}^t
    =
    \sum_{k=1}^{K} \alpha_k U_k^t.
\end{equation}

Since full participation is often infeasible under wireless resource constraints, the server selects a subset $\mathcal{A}^t$ of clients. Let $p_k^t$ denote the probability that client $k$ participates in round $t$. The sampled update is aggregated using inverse-probability weighting:
\begin{equation}
    \widehat{U}^t
    =
    \sum_{k \in \mathcal{A}^t}
    \frac{\alpha_k}{p_k^t} U_k^t.
\end{equation}
This gives an unbiased estimator of the full-participation update:
\begin{align}
    \mathbb{E}_{\mathcal{A}^t} \left[\widehat{U}^t \middle|\{U_k^t\}_{k=1}^{K}\right] =
    \sum_{k=1}^{K} \frac{\mathbb{E}\left[\mathbf{1}_{k\in\mathcal{A}^t}\right]\alpha_k U_k^t}{p_k^t} =
    \sum_{k=1}^{K} \alpha_k U_k^t.
    \label{eq:fullparticipation}
\end{align}

\subsection{Variance-Minimizing Sampling Objective}

Since only a subset of clients participates in each training round, the aggregated update becomes a random estimate of the full-participation update \cite{zhang2025towards}. To improve the quality of this estimate, optimal client sampling chooses the participation probabilities $p_k^t$ so as to minimize the variance of the sampled update. Lower variance means that the update obtained from sampled clients remains closer to the update that would be obtained if all clients participated, leading to more stable training. This variance is defined as
\begin{equation}
\mathcal{V}^t(p^t)
=
\mathbb{E}_{\mathcal{A}^t}
\left[
\left\|
\sum_{k \in \mathcal{A}^t}
\frac{\alpha_k}{p_k^t} U_k^t
-
\sum_{k=1}^{K} \alpha_k U_k^t
\right\|^2
\right].
\label{eq:variance_objective}
\end{equation}

Under independent Bernoulli sampling, client $k$ is selected independently with probability $p_k^t$. In this case, $\widehat{U}^t$ is unbiased. By expanding \eqref{eq:variance_objective}, the cross terms vanish due to independence, and using
$\mathrm{Var}(\mathbf{1}_{k\in\mathcal{A}^t})=p_k^t(1-p_k^t)$, the variance admits the closed form
\begin{equation}
\mathcal{V}^t(p^t)
=
\sum_{k=1}^{K}
\alpha_k^2
\frac{1-p_k^t}{p_k^t}
\|U_k^t\|^2.
\label{eq:variance_closed_form}
\end{equation}
\RAOCSchanged{A detailed derivation of \eqref{eq:variance_closed_form} is provided in Appendix~\ref{app:variance_derivation}.}

This objective shows that sampling depends on the update-importance term $\|U_k^t\|$, which motivates the update-information model introduced next.

\subsection{Update-Information Availability Model}
\label{sub:update-information}

\RAOCSchanged{The variance objective in \eqref{eq:variance_objective} depends on the current update norms $\|U_k^t\|$. Obtaining these norms from every client before each sampling decision would require all clients to perform local training, even though only a subset will eventually upload their updates. This incurs additional computation costs, particularly in heterogeneous or energy-constrained systems.}

\RAOCSchanged{To address this limitation, the server maintains a cached update-information estimate $\widetilde{U}_k^t = U_k^{t-s_k}$ for each client $k$, where $s_k$ denotes the number of rounds since its last update computation. The corresponding norm $\|\widetilde{U}_k^t\|$ is available at the beginning of round $t$, but may become less representative of the current norm $\|U_k^t\|$ as $s_k$ increases.}

\RAOCSchanged{To mitigate this staleness, we introduce probabilistic information refreshing. After the participation decision in round $t$, each non-selected client independently computes a fresh local update using $w^t$ with probability $q\in[0,1]$, solely to renew its cached information. Selected clients already compute fresh updates and therefore renew their information as well. A refreshed estimate, $\widetilde{U}_k^{t+1}=U_k^t$, becomes available for the following round, while clients that neither participate nor refresh retain their previous estimates. Hence, $q$ controls the trade-off between information freshness for subsequent sampling decisions and additional local computation costs.}

\subsection{Wireless Resource-Block Communication Model}

We consider a wireless FL system where client updates are uploaded through a limited number of uplink resource blocks (RBs). At round $t$, the system has a total resource-block budget $B$. If client $k$ participates, it is assigned $b_k^t$ effective uplink RBs. Since client $k$ participates with probability $p_k^t$, we impose the expected resource-block constraint
\begin{equation}
\sum_{k=1}^{K} p_k^t b_k^t \leq B,
\qquad b_k^t \in \mathbb{R}^{+}.
\label{eq:rb_constraint}
\end{equation}
Although $B$ is the number of available physical RBs, $b_k^t$ is modeled as an effective average allocation and may be fractional, since RBs can be scheduled over short time intervals during a multi-second upload.

The wireless channel is heterogeneous across clients and varies across rounds. We denote by $\nu_k^t$ the uplink rate provided by one RB to client $k$ at round $t$, modeled as
\begin{equation}
\nu_k^t
=
\xi W_{\mathrm{RB}}
\log_2\left(
1+
\frac{P_k^t |h_k^t|^2}
{\sigma_k^{2,t}\ell_k^t}
\right),
\label{eq:uplink_rate}
\end{equation}
where $W_{\mathrm{RB}}$ is the bandwidth of each RB, $\xi\in(0,1]$ is a rate-efficiency factor accounting for practical coding and protocols, $P_k^t$ is the uplink transmit power of client $k$, $h_k^t$ is the channel coefficient, $\sigma_k^{2,t}$ is the noise power, and $\ell_k^t$ captures large-scale attenuation such as path loss and shadowing. Thus, a client assigned $b_k^t$ RBs has an effective uplink rate $\nu_k^t b_k^t$.

When client $k$ is selected, it uploads a fresh model update of size $Z$ bits to the server through its assigned RBs. Its uplink transmission time at round $t$ is
\begin{equation}
    T_k^{\uparrow,t}
    =
    \frac{Z}{\nu_k^t b_k^t}.
    \label{eq:upload_time}
\end{equation}

Let $\tau$ be the round deadline and let $T_k^{\mathrm{comp},t}$ be the local computation time required to produce a fresh update. \RAOCSchanged{We assume that $T_k^{\mathrm{comp},t}$ is known to the server before the sampling and resource-allocation decisions.} The round-deadline constraint is
\begin{equation}
\frac{Z}{\nu_k^t b_k^t}
+
T_k^{\mathrm{comp},t}
\leq
\tau.
\label{eq:deadline_constraint}
\end{equation}
From \eqref{eq:deadline_constraint}, the minimum effective number of RBs required by client $k$ is
\begin{equation}
b_{k,\min}^t
=
\frac{Z}
{\nu_k^t\left(\tau-T_k^{\mathrm{comp},t}\right)}.
\label{eq:min_blocks}
\end{equation}

Clients with higher uplink rates require fewer RBs to meet the deadline, whereas clients with poor channels require more resources and are costly to schedule. Consequently, the deadline $\tau$ is a critical parameter. We assume that $\tau$ is chosen so that each client can meet the deadline when allocated sufficient bandwidth. Later, the effect of computation-time variability and client availability is studied in Section~\ref{sub:comp_time_sensitivity}.

\subsection{Energy Model}

In addition to uplink resource constraints, clients may be subject to limited energy budgets. Let $E_k$ denote the maximum expected energy available to client $k$ for FL per round. This budget accounts for the energy required to compute a fresh local update and the energy required to upload it when the client is selected.

Let $\zeta_k^{\mathrm{comp},t}$ and $\zeta_k^{\uparrow,t}$ denote the computation energy and uplink energy of client $k$ at round $t$, respectively:
\begin{equation}
    \zeta_k^{\uparrow,t}
    =
    P_k^t T_k^{\uparrow,t},
    \qquad
    \zeta_k^{\mathrm{comp},t}
    =
    \kappa_k i_k^t \beta_k c_k (f_k^t)^2,
    \label{eq:energy_components}
\end{equation}
where $P_k^t$ is the uplink transmit power in \eqref{eq:uplink_rate}, $T_k^{\uparrow,t}$ is the uplink transmission time in \eqref{eq:upload_time}, $\kappa_k$ is the effective switched capacitance, $f_k^t$ is the CPU/GPU frequency, $i_k^t$ is the number of local iterations, $\beta_k$ is the batch size, and $c_k$ is the number of cycles required to process one training sample.

\RAOCSchanged{Under the update-information model, a selected client computes a fresh update and incurs both computation and uplink energy. A non-selected client incurs computation energy only if it refreshes its stored information, which occurs with probability $q$. Hence, the expected energy constraint is}
\begin{equation}
    p_k^t
    \left(
    \zeta_k^{\mathrm{comp},t}
    +
    \zeta_k^{\uparrow,t}
    \right)
    +
    (1-p_k^t)q\zeta_k^{\mathrm{comp},t}
    \leq
    E_k.
    \label{eq:energy_constraint}
\end{equation}

Equivalently, \eqref{eq:energy_constraint} imposes the following upper bound on the sampling probability:
\begin{equation}
    p_k^t
    \leq
    \widehat{p}_k^t(q)
    =
    \frac{
    E_k - q\zeta_k^{\mathrm{comp},t}
    }{
    (1-q)\zeta_k^{\mathrm{comp},t}
    +
    \zeta_k^{\uparrow,t}
    }.
    \label{eq:energy_probability_bound}
\end{equation}

We assume that $q$ is chosen such that
$E_k \geq q\zeta_k^{\mathrm{comp},t}$ for each considered client, so that the refresh energy remains feasible. The resulting probability cap is
\begin{equation}
    \bar{p}_k^t
    =
    \min
    \left\{
    1,
    \widehat{p}_k^t(q)
    \right\}.
    \label{eq:probability_cap}
\end{equation}

The cap depends on both the energy budget and the refresh probability. Larger values of $q$ increase \RAOCSchanged{information-refresh computation} energy and make the cap more restrictive. Smaller energy budgets also impose stricter caps, whereas sufficiently large budgets make the cap inactive, i.e., $\bar{p}_k^t=1$.

\subsection{Resource-Constrained OCS Problem}

Combining the sampling-variance objective in \eqref{eq:variance_closed_form} with the resource-block constraint in \eqref{eq:rb_constraint} and the energy-induced probability cap in \eqref{eq:probability_cap}, the proposed resource-constrained OCS problem at round $t$ is
\begin{align}
    \min_{p^t} \quad &
    \sum_{k=1}^{K}
    \alpha_k^2
    \frac{1-p_k^t}{p_k^t}
    \|U_k^t\|^2
    \label{eq:main_problem} \\
    \text{s.t.} \quad &
    \sum_{k=1}^{K} p_k^t b_{k,\min}^t \leq B,
    \label{eq:expected_rb_constraint} \\
    & 0 < p_k^t \leq \bar{p}_k^t,
    \quad \forall k \in \mathcal{K}.
\end{align}

This formulation preserves the variance-minimization objective of OCS while replacing the client-count budget with explicit wireless and energy constraints. The term $\|U_k^t\|$ captures the statistical importance of client $k$, $b_{k,\min}^t$ captures its uplink resource cost required to satisfy the deadline, and $\bar p_k^t$ limits its participation according to the energy budget and refresh probability. Hence, the sampling distribution balances update importance, wireless feasibility, and energy availability. Unlike conventional wireless scheduling, which favors clients with good channels, this objective can still assign high participation probabilities to poor-channel clients when their updates are statistically important.

\section{Optimal Solution}

\subsection{Capped Water-Filling Solution}
At each round $t$, the server solves the resource-constrained OCS problem in \eqref{eq:main_problem}. The objective is convex with respect to $p_k^t$ for $p_k^t>0$, and the constraints are linear.

\begin{theorem}
The closed-form optimal solution is given by
\begin{equation}
p_k^{t,\star}
=
\begin{cases}
\displaystyle
\frac{
B-\sum_{j\notin\mathcal{O}^t} \bar{p}_j^t b_{j,\min}^t
}{
\sum_{j\in\mathcal{O}^t}
\sqrt{b_{j,\min}^t}\,\alpha_j\|U_j^t\|
}
\cdot
\frac{\alpha_k\|U_k^t\|}
{\sqrt{b_{k,\min}^t}},
& k\in\mathcal{O}^t, \\[3ex]
\bar{p}_k^t,
& k\notin\mathcal{O}^t,
\end{cases}
\label{eq:water_filling_piecewise}
\end{equation}
where $\mathcal{O}^t$ is the set of non-saturated clients, characterized by
\begin{equation}
\mathcal{O}^t
=
\left\{
k\in\mathcal{K}
\; \middle| \;
\frac{\alpha_k\|U_k^t\|}
{\bar{p}_k^t\sqrt{b_{k,\min}^t}}
<
\frac{
\sum_{j\in\mathcal{O}^t}
\sqrt{b_{j,\min}^t}\,\alpha_j\|U_j^t\|
}{
B-\sum_{j\notin\mathcal{O}^t} \bar{p}_j^t b_{j,\min}^t
}
\right\}.
\label{eq:ocs_resource_set}
\end{equation}
\end{theorem}

The solution follows from the KKT conditions; the derivation is provided in Appendix~\ref{app:kkt_solution}. It has a capped water-filling structure: clients saturate according to the threshold
$\frac{\bar{p}_k^t\sqrt{b_{k,\min}^t}}{\alpha_k\|U_k^t\|}$.
Clients with smaller thresholds reach their maximum feasible probabilities $\bar{p}_k^t$ first, while the remaining clients form $\mathcal{O}^t$ and receive probabilities according to \eqref{eq:water_filling_piecewise}.

The rule balances statistical importance, wireless cost, and energy feasibility. Larger weighted update norms increase the sampling probability, whereas larger resource-block requirements or restrictive energy caps reduce it.

\RAOCSchanged{The optimal rule in \eqref{eq:water_filling_piecewise} involves the current update norm $\|U_k^t\|$, which is generally unknown before participation is decided. In practice, RAOCS evaluates the rule using the stored norm $g_k^t=\|\widetilde{U}_k^t\|$ available at the beginning of round $t$ instead of $\|U_k^t\|$. This substitution uses only information already available at the sampling decision.}

\subsection{RAOCS Algorithm}

\begin{algorithm}[h]
\caption{Resource-Aware Optimal Client Sampling}
\label{alg:raocs}
\begin{algorithmic}[1]
\REQUIRE Resource budget $B$, upload deadline $\tau$, round size $Z$, refresh probability $q$, initial model $w^0$
\STATE \RAOCSchanged{Initialize stored update norms $g_k^0$ for all clients.}
\FOR{each round $t=0,1,\ldots$}
\STATE Broadcast $w^t$ to clients.
\STATE Collect or estimate uplink rates $\nu_k^t$.
\STATE Compute $b_{k,\min}^t$ using \eqref{eq:min_blocks}.
\STATE Compute the probability caps $\bar{p}_k^t$ using \eqref{eq:probability_cap}.
\STATE \RAOCSchanged{Collect latest stored norms $g_k^t$ from clients.}
\STATE \RAOCSchanged{Compute $p_k^t$ using \eqref{eq:water_filling_piecewise} with $g_k^t$ in place of $\|U_k^t\|$.}
\STATE Sample the participating set $\mathcal{A}^t$ according to $p^t$.
\STATE Allocate $b_{k,\min}^t$ uplink resource blocks to each $k\in\mathcal{A}^t$.
\FORALL{$k \in \mathcal{K}$ \textbf{in parallel}}
    \IF{$k \in \mathcal{A}^t$}
        \STATE \RAOCSchanged{Compute the fresh local update $U_k^t$ using $w^t$.}
        \STATE \RAOCSchanged{Upload $U_k^t$ to the server.}
        \STATE \RAOCSchanged{Set $g_k^{t+1} \gets \|U_k^t\|$.}
    \ELSE
        \STATE \RAOCSchanged{Draw $z_k^t \sim \mathrm{Bernoulli}(q)$.}
        \IF{$z_k^t = 1$}
            \STATE \RAOCSchanged{Compute a fresh local update $U_k^t$ using $w^t$.}
            \STATE \RAOCSchanged{Set $g_k^{t+1} \gets \|U_k^t\|$.}
        \ELSE
            \STATE \RAOCSchanged{Set $g_k^{t+1} \gets g_k^t$.}
        \ENDIF
    \ENDIF
\ENDFOR
\STATE Aggregate $\widehat{U}^t = \sum_{k\in\mathcal{A}^t}
    \frac{\alpha_k}{p_k^t}U_k^t$.
\STATE Update $w^{t+1}=w^t+\widehat{U}^t$.
\ENDFOR
\end{algorithmic}
\end{algorithm}

\begin{figure}[h]
\centering
\includegraphics[width=\columnwidth]{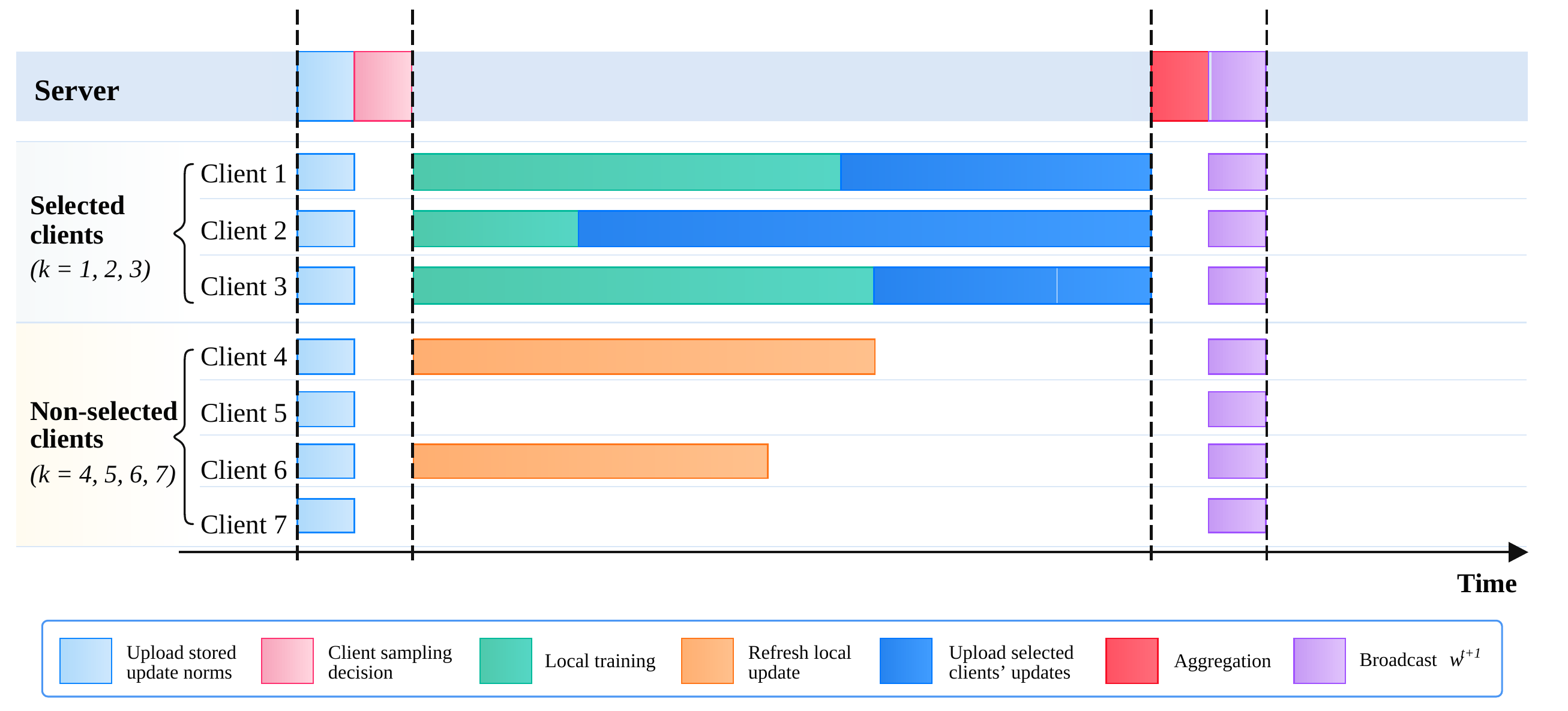}
\caption{\RAOCSchanged{Illustration of RAOCS Algorithm~\ref{alg:raocs} during round $t$. Clients report their stored update norms (light blue), and the server selects participants using this information (pink). Selected clients compute (turquoise) and upload (blue) fresh updates, while non-selected clients may compute local updates solely to refresh their stored norms (orange). Finally, the server aggregates the received updates to obtain $w^{t+1}$ (red) and broadcasts the new model to all clients (purple).}}
\label{fig:raocs_timeline}
\end{figure}

Algorithm~\ref{alg:raocs} summarizes the proposed RAOCS framework. \RAOCSchanged{The server first determines the sampling probabilities using the stored norms and then selects the participating clients. Selected clients train using the current global model and upload fresh updates for aggregation. Non-selected clients may train with probability $q$ only to renew their stored update norms, without uploading full updates. All norms refreshed during round $t$ are used from round $t+1$ onward. The round chronology is illustrated in Fig.~\ref{fig:raocs_timeline}.}

\section{Experiments}

\subsection{Experimental Setup}

We evaluate RAOCS in simulated wireless FL settings with $K=50$ clients. Data are partitioned across clients using a Dirichlet distribution with $\alpha_{\mathrm{Dir}}=0.2$ to model non-IID heterogeneity. Each selected client performs 16 local SGD steps per round, using mini-batches of size 64 and learning rate $0.01$.

We consider three image-classification settings: Fashion-MNIST with a custom fully connected network\footnote{4 FC layers of sizes: $784\to256\to128\to64\to10$}, CIFAR-10 with ResNet-8, and CIFAR-100 with ResNet-20. For each setting, all methods use the same data partition, model initialization protocol, and wireless trace.


The wireless trace is generated by a simulator where clients move inside a circular cell around a base station. The per-resource-block uplink rate $\nu_k^t$ in \eqref{eq:uplink_rate} varies across clients and rounds due to mobility, path loss, shadowing, Rayleigh fading, transmit power, and noise.

Unless otherwise stated, local computation time is assumed identical across clients and methods and is absorbed into the round deadline $\tau$. Thus, $\tau$ denotes the remaining time available for uplink transmission, and the communication setting is controlled by $\tau$ and the resource-block budget $B$. We also use the \RAOCSchanged{full-refresh} configuration $q=1$ with inactive energy caps, i.e., $\bar p_k^t=1$, unless stated otherwise.

\subsection{Baseline Comparison}

\begin{figure*}[t]
    \centering

    \subfloat{
        \includegraphics[width=0.31\textwidth]{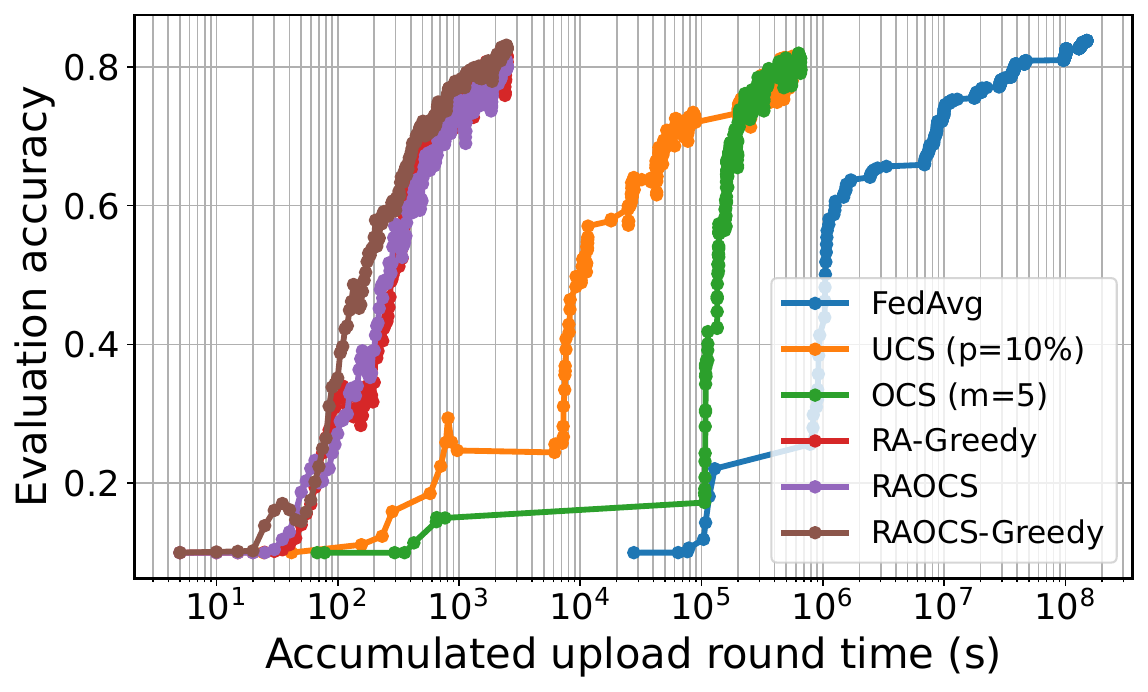}
        \label{fig:FCN-MNIST}
    }
    \hfill
    \subfloat{
        \includegraphics[width=0.31\textwidth]{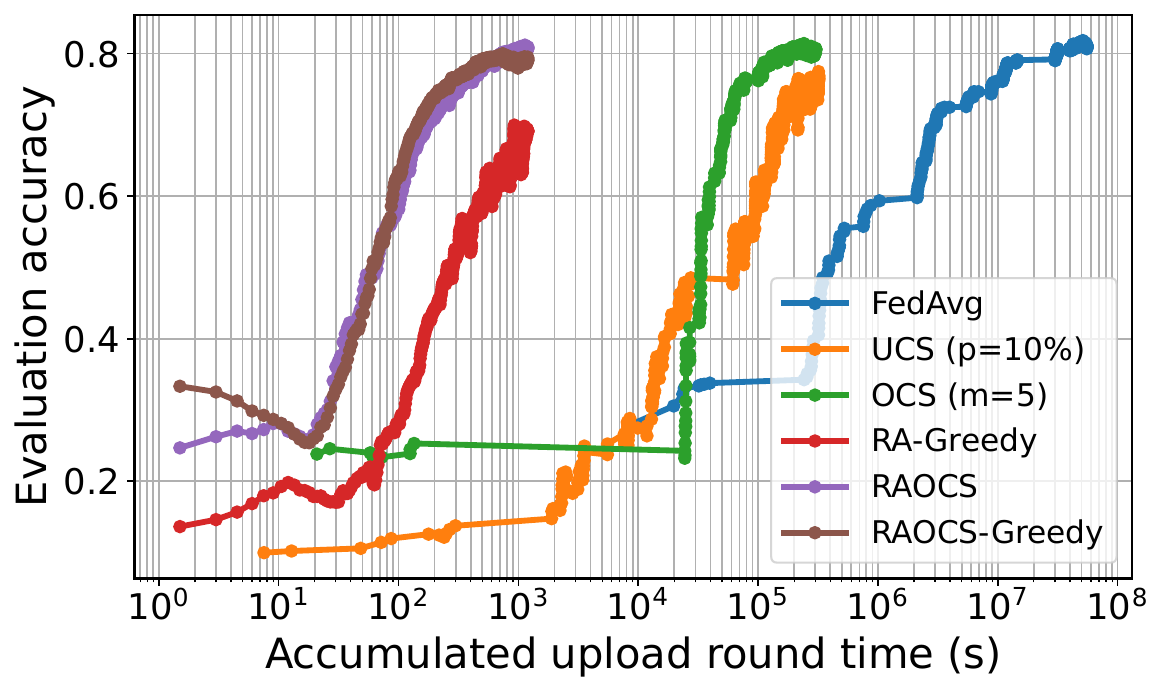}
        \label{fig:ResNet8-CIFAR10}
    }
    \hfill
    \subfloat{
        \includegraphics[width=0.31\textwidth]{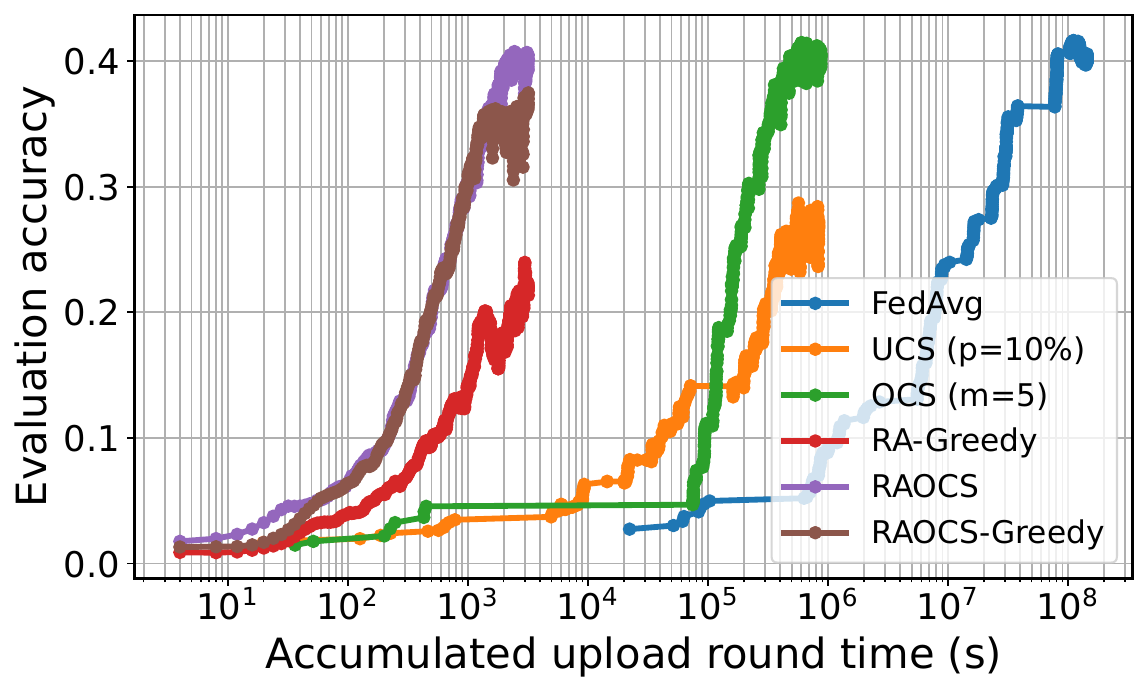}
        \label{fig:ResNet20-CIFAR100}
    }

    \caption{Evaluation accuracy versus accumulated update upload time for the compared methods under three dataset/model settings: Fashion-MNIST/Custom-FCN, CIFAR-10/ResNet-8, and CIFAR-100/ResNet-20.}
    
    \label{fig:baseline-accuracy-vs-time}
\end{figure*}

We first compare RAOCS with sampling and resource-aware baselines: FedAvg, UCS, OCS, RA-Greedy, and RAOCS-Greedy. FedAvg is the full-participation reference. UCS samples clients uniformly with probability $p_{UCS}$, while OCS performs variance-aware sampling under a client-count budget $m$ but ignores wireless constraints. RA-Greedy selects clients with the smallest resource-block requirements but doesn't consider the sampling variance. RAOCS-Greedy is the strict-budget counterpart of RAOCS: it ranks clients by a score $s_k^t=\frac{\alpha_k|U_k^t|}{\sqrt{b_{k,\min}^t}}$, inspired from the optimal solution in \eqref{eq:water_filling_piecewise}, and selects clients in decreasing order while ensuring that the realized resource usage satisfies $\sum_{k\in A^t} b_{k,\min}^t \leq B$.

For fairness, UCS and OCS use the average number of clients selected by RAOCS, denoted by $\bar m$, under the same wireless setting. Thus, UCS uses $p_{\mathrm{ucs}}=\bar m/K$, and OCS uses a client-count budget $m=\bar m$. For FedAvg, UCS, and OCS, which do not perform resource-aware allocation, the available resource blocks are divided equally among selected clients.

We evaluate the methods on the three dataset/model pairs: Fashion-MNIST with a custom FCN, CIFAR-10 with ResNet-8, and CIFAR-100 with ResNet-20. For these settings, we use $(B=5,\tau=5)$, $(B=5,\tau=1.5)$, and $(B=7,\tau=4)$, respectively. We run 500 communication rounds on Fashion-MNIST and 800 rounds on CIFAR-10 and CIFAR-100.

For each method, we report test accuracy versus accumulated update upload time in Fig.~\ref{fig:baseline-accuracy-vs-time}. We also report in Table~\ref{tab:baseline-stats} the final accuracy, average realized resource-block usage $\bar{B}$, average round upload time $\bar{T}^{\uparrow}$, and average participation $\bar{m}$ over all rounds, with their standard deviations $\sigma_B$, $\sigma_T$, and $\sigma_m$.

\begin{table}[h]
\centering
\caption{Baseline comparison in terms of accuracy, resource block usage, upload time, and client participation.}
\label{tab:baseline-stats}
\scriptsize

\resizebox{\columnwidth}{!}{%
\begin{tabular}{lcccc}
\toprule
\multicolumn{5}{c}{FCN model + Fashion MNIST} \\
\midrule
& Acc. & $\bar{B}\pm\sigma_B$ & $\bar{T}^{\uparrow}\pm\sigma_T$ (s) & $\bar{m}\pm\sigma_m$ \\
\midrule

FedAvg
& 83.73 & 5 & 301705 $\pm$ 2552324 & 50 \\

UCS (p=10\%)
& 79.95 & 5 & 1161 $\pm$ 5980 & 5 \\

OCS (m=5)
& 79.51 & 5 & 1318 $\pm$ 6302 & 5.03 $\pm$ 2.15 \\

RA-Greedy
& 81.41 & \textbf{4.65 $\pm$ 0.2} & \textbf{5} & \textbf{10.28 $\pm$ 1.04} \\

RAOCS-Greedy
& \textbf{82.58} & \textbf{4.86 $\pm$ 0.11} & \textbf{5} & 8.23 $\pm$ 1.34 \\

RAOCS
& 80.15 & 5.17 $\pm$ 2.93 & \textbf{5} & 4.72 $\pm$ 2.05 \\

\bottomrule
\end{tabular}%
}

\vspace{0.7em}

\resizebox{\columnwidth}{!}{%
\begin{tabular}{lcccc}
\toprule
\multicolumn{5}{c}{ResNet-8 + CIFAR-10} \\
\midrule
& Acc. & $\bar{B}\pm\sigma_B$ & $\bar{T}^{\uparrow}\pm\sigma_T$ (s) & $\bar{m}\pm\sigma_m$ \\
\midrule

FedAvg
& 81.03 & 5 & 69225 $\pm$ 623419 & 50 \\

UCS (p=10\%)
& 76.86 & 5 & 402 $\pm$ 1975 & 5 \\

OCS (m=5)
& 80.62 & 5 & 377 $\pm$ 1615 & 4.97 $\pm$ 2.08 \\

RA-Greedy
& 69.11 & \textbf{4.64 $\pm$ 0.22} & \textbf{1.5} & \textbf{10.18 $\pm$ 1.05} \\

RAOCS-Greedy
& 79.02 & \textbf{4.85 $\pm$ 0.12} & \textbf{1.5} & 7.90 $\pm$ 1.62 \\

RAOCS
& \textbf{80.95} & 5.10 $\pm$ 2.93 & \textbf{1.5} & 4.83 $\pm$ 1.99 \\

\bottomrule
\end{tabular}%
}

\vspace{0.7em}

\resizebox{\columnwidth}{!}{%
\begin{tabular}{lcccc}
\toprule
\multicolumn{5}{c}{ResNet-20 + CIFAR-100} \\
\midrule
& Acc. & $\bar{B}\pm\sigma_B$ & $\bar{T}^{\uparrow}\pm\sigma_T$ (s) & $\bar{m}\pm\sigma_m$ \\
\midrule

FedAvg
& 40.27 & 7 & 181233 $\pm$ 1632126 & 50 \\

UCS (p=10\%)
& 27.24 & 7 & 1051 $\pm$ 5172 & 5 \\

OCS (m=5)
& 39.53 & 7 & 1094 $\pm$ 4779 & 5 $\pm$ 2.1 \\

RA-Greedy
& 21.35 & \textbf{6.52 $\pm$ 0.29} & \textbf{4} & \textbf{10.33 $\pm$ 1.05} \\

RAOCS-Greedy
& 37.46 & \textbf{6.70 $\pm$ 0.23} & \textbf{4} & 9.5 $\pm$ 1.31 \\

RAOCS
& \textbf{40.10} & 7.06 $\pm$ 4.27 & \textbf{4} & 4.62 $\pm$ 2.05 \\

\bottomrule
\end{tabular}%
}

\end{table}

FedAvg provides the full-participation reference, but is highly inefficient in wireless time because all clients upload every round. UCS reduces participation, but ignores both update importance and wireless cost, leading to lower accuracy, especially on CIFAR-100.

OCS remains statistically strong because it samples clients according to update importance, reaching accuracies close to FedAvg on CIFAR-10 and CIFAR-100. However, it is not wireless-aware, so selected clients may require many resource blocks and long upload times. On CIFAR-10, OCS reaches 80.62\% accuracy but requires 377 s per round on average, compared with 80.95\% and 1.5 s for RAOCS. On CIFAR-100, OCS reaches 39.53\% with 1094 s per round, while RAOCS reaches 40.10\% with 4 s.

The naive RA-Greedy baseline shows that minimizing wireless cost alone is insufficient. Although it selects around 10 clients per round and satisfies the deadline constraint, it ignores update importance, reaching only 69.11\% on CIFAR-10 and 21.35\% on CIFAR-100. Thus, selecting more clients does not necessarily improve learning when the selected updates are not statistically informative.

RAOCS balances update importance and resource cost. It preserves the variance-aware objective of OCS while penalizing clients with high resource-block requirements, achieving OCS-level accuracy with deadline-limited upload time. Its larger RB standard deviation reflects its probabilistic design, where the budget is enforced in expectation. RAOCS-Greedy is the strict-budget counterpart: it keeps realized RB usage below the budget with small variability and improves over RA-Greedy by using the resource-aware OCS score instead of resource cost alone. Overall, RAOCS is preferable when expected feasibility is acceptable, whereas RAOCS-Greedy is preferable when the RB budget must be strictly respected each round.

\subsection{Sensitivity to Upload Deadline}

We study the effect of the upload deadline $\tau$ on RAOCS. We fix the resource-block budget $B$ while varying $\tau$ to model stricter or more relaxed uplink latency requirements. The study is conducted on CIFAR-10 with ResNet-8 and CIFAR-100 with ResNet-20. Table~\ref{tab:tau_sensitivity} reports the final accuracy, accumulated upload time, and average number of selected clients along with its standard deviation.

\begin{table}[h]
\centering
\caption{Sensitivity of RAOCS to the upload deadline $\tau$.}
\label{tab:tau_sensitivity}

\resizebox{\columnwidth}{!}{%
\begin{tabular}{cccc}
\toprule
\multicolumn{4}{c}{ResNet-8 + CIFAR-10} \\
\midrule
$\tau$ & Final Acc. & Accumulated upload time (s) & Selected clients \\
\midrule
0.2 & 77.24 & 160  & 1.76 $\pm$ 0.92 \\
0.6 & 79.41 & 480  & 2.82 $\pm$ 1.46 \\
1.5 & 80.95 & 1200 & 4.83 $\pm$ 1.99 \\
3.0 & 81.22 & 2400 & 7.42 $\pm$ 2.50 \\
6.0 & 81.00 & 4800 & 11.55 $\pm$ 2.94 \\
\bottomrule
\end{tabular}
}

\vspace{0.7em}

\resizebox{\columnwidth}{!}{%
\begin{tabular}{cccc}
\toprule
\multicolumn{4}{c}{ResNet-20 + CIFAR-100} \\
\midrule
$\tau$ & Final Acc. & Accumulated upload time (s) & Selected clients \\
\midrule
0.5 & 32.63 & 400  & 1.71 $\pm$ 0.95 \\
1.0 & 36.74 & 800  & 2.19 $\pm$ 1.26 \\
2.0 & 39.27 & 1600 & 3.06 $\pm$ 1.65 \\
4.0 & 40.10 & 3200 & 4.62 $\pm$ 2.05 \\
8.0 & 40.02 & 6400 & 7.21 $\pm$ 2.63 \\
\bottomrule
\end{tabular}
}

\end{table}

Increasing $\tau$ relaxes the uplink constraint because each selected client requires fewer resource blocks to meet the deadline. As a result, RAOCS selects more clients per round. On CIFAR-10, participation increases from $1.76$ clients at $\tau=0.2$ to $11.55$ at $\tau=6.0$; on CIFAR-100, it increases from $1.71$ at $\tau=0.5$ to $7.21$ at $\tau=8.0$. This improves accuracy under strict deadlines, from $77.24\%$ at $\tau=0.2$ to $81.22\%$ at $\tau=3.0$ on CIFAR-10 and from $32.63\%$ at $\tau=0.5$ to $40.10\%$ at $\tau=4.0$ on CIFAR-100.

The gain saturates after a moderate deadline. On CIFAR-10, increasing $\tau$ from $3.0$ to $6.0$ doubles the accumulated upload time from $2400$ s to $4800$ s, while accuracy slightly decreases from $81.22\%$ to $81.00\%$. Similarly, on CIFAR-100, increasing $\tau$ from $4.0$ to $8.0$ doubles the accumulated upload time from $3200$ s to $6400$ s, while accuracy remains almost unchanged. Thus, $\tau$ controls the trade-off between participation, accuracy, and uplink time, with moderate deadlines providing the best balance.

\subsection{Sensitivity to Resource-Block Availability}

We evaluate the robustness of RAOCS and RAOCS-Greedy when the uplink resource-block budget varies across rounds. This reflects practical settings where the resources available for FL may change due to background traffic, network load, or dynamic resource sharing with other services, while keeping the same average budget.

To model this variability, at each round $t$ we sample an integer budget $B^t$ from a rounded Gaussian distribution with mean $\mu_B=5$ and standard deviation $\sigma_B$, lower bounded by one resource block. The sampled value $B^t$ replaces the fixed budget in both RAOCS and RAOCS-Greedy.

\begin{table}[h]
\centering
\caption{Sensitivity to time-varying resource-block availability on CIFAR-10 with ResNet-8.}
\label{tab:variable_B_sensitivity}
\begin{tabular}{lccc}
\toprule
Method & $\sigma_B$ & Acc. (\%) & Selected Clients \\
\midrule
\multirow{4}{*}{RAOCS}
& 0 & 80.95 & $4.83 \pm 1.99$ \\
& 1 & 80.38 & $4.60 \pm 2.00$ \\
& 2 & 80.71 & $4.54 \pm 2.21$ \\
& 4 & 80.93 & $4.56 \pm 2.63$ \\
\midrule
\multirow{4}{*}{RAOCS-Greedy}
& 0 & 79.89 & $7.80 \pm 1.54$ \\
& 1 & 79.16 & $7.75 \pm 1.90$ \\
& 2 & 79.61 & $7.62 \pm 2.62$ \\
& 4 & 79.50 & $7.51 \pm 3.96$ \\
\bottomrule
\end{tabular}
\end{table}

Table~\ref{tab:variable_B_sensitivity} shows that both methods are robust to time-varying resource availability. RAOCS remains around $80.75\%$ accuracy for all values of $\sigma_B$, while RAOCS-Greedy stays close to $79.5\%$. The average number of selected clients changes only slightly as $\sigma_B$ increases, but its standard deviation grows, indicating that larger budget fluctuations mainly increase round-to-round variability in participation. Overall, both methods remain stable under resource-budget variability.

\subsection{Effect of Energy Caps and Update-Information}

We evaluate how the refresh probability $q$ and the energy-induced probability caps $\bar p_k^t$ affect RAOCS. We use CIFAR-10 with ResNet-8 under $B=10$ and $\tau=6$ s, and CIFAR-100 with ResNet-20 under $B=30$ and $\tau=16$ s. For each dataset, we compare four configurations: full update-information refresh $(q=1)$ with no energy cap $(\bar p_k^t=1)$, full refresh with energy caps $(\bar p_k^t<1)$, cached update information $(q=0)$ with no energy cap, and cached update information with energy caps. \RAOCSchanged{For $q=1$, all clients renew their norms after the current sampling decision for use in the next round; for $q=0$, only selected clients renew their norms.} All configurations use the same data partition, wireless trace, and energy profiles.

Table~\ref{tab:energy_extension} reports the accumulated upload time and accumulated energy consumption at the first round where target accuracies are reached. We use target accuracies of $65\%$, $75\%$, and $80\%$ for CIFAR-10, and $30\%$, $35\%$, and $38\%$ for CIFAR-100.

\begin{table}[h]
\centering
\caption{Effect of update-information refresh and energy caps at multiple target accuracies. Time is reported in seconds and energy in kJ.}
\label{tab:energy_extension}

\resizebox{\columnwidth}{!}{%
\begin{tabular}{lcccccc}
\toprule
\multicolumn{7}{c}{ResNet-8 + CIFAR-10} \\
\midrule
Configuration & \multicolumn{2}{c}{Target $65\%$} & \multicolumn{2}{c}{Target $75\%$} & \multicolumn{2}{c}{Target $80\%$} \\
\cmidrule(lr){2-3}\cmidrule(lr){4-5}\cmidrule(lr){6-7}
& Time & Energy & Time & Energy & Time & Energy \\
\midrule
$q=1$, no cap & 414 & 165.9 & 900 & 356.7 & 1908 & 752.3 \\
$q=1$, energy cap & 414 & 139.6 & 948 & 316.3 & 2040 & 682.1 \\
$q=0$, no cap & 714 & 137.3 & 1308 & 257.5 & 2628 & 507.2 \\
$q=0$, energy cap & 720 & 131.1 & 1350 & 252.2 & 2472 & 453.1 \\
\bottomrule
\end{tabular}%
}

\vspace{0.7em}

\resizebox{\columnwidth}{!}{%
\begin{tabular}{lcccccc}
\toprule
\multicolumn{7}{c}{ResNet-20 + CIFAR-100} \\
\midrule
Configuration & \multicolumn{2}{c}{Target $30\%$} & \multicolumn{2}{c}{Target $35\%$} & \multicolumn{2}{c}{Target $38\%$} \\
\cmidrule(lr){2-3}\cmidrule(lr){4-5}\cmidrule(lr){6-7}
& Time & Energy & Time & Energy & Time & Energy \\
\midrule
$q=1$, no cap & 2624 & 1789.3 & 3536 & 2412.8 & 4448 & 3028.1 \\
$q=1$, energy cap & 2800 & 1452.3 & 4240 & 2196.2 & 4928 & 2557.5 \\
$q=0$, no cap & 3296 & 1685.7 & 4448 & 2237.1 & 5584 & 2785.8 \\
$q=0$, energy cap & 3488 & 1554.0 & 4832 & 2116.4 & 5472 & 2396.5 \\
\bottomrule
\end{tabular}%
}

\end{table}

Table~\ref{tab:energy_extension} shows two effects. First, for a fixed refresh setting $q$, energy caps reduce the energy needed to reach the target accuracies. With $q=1$, energy decreases by $9.3$-$15.9\%$ on CIFAR-10 and by $9.0$-$18.8\%$ on CIFAR-100. With $q=0$, the reductions are $2.1$-$10.7\%$ on CIFAR-10 and $5.4$-$14.0\%$ on CIFAR-100. This confirms that the probability caps effectively limit energy-expensive participation across both datasets.

Second, reducing the refresh probability from $q=1$ to $q=0$ lowers computation energy because \RAOCSchanged{non-selected clients do not perform additional local training to refresh stored norms}. This is visible even without energy caps: on CIFAR-10, energy drops from $752.3$ kJ to $507.2$ kJ at the $80\%$ target. However, using cached information can increase the time-to-target because the sampling probabilities are computed from less fresh update information. For example, on CIFAR-10 without caps, the time to reach $80\%$ increases from $1908$ s with $q=1$ to $2628$ s with $q=0$.

Overall, $q$ and $\bar p_k^t$ control different parts of the energy-time trade-off. The refresh probability $q$ controls \RAOCSchanged{the computation cost of refreshing stored update-norm information}, while the cap $\bar p_k^t$ limits the participation of energy-expensive clients. Combining cached information with energy caps gives the lowest energy at the highest target on both datasets: $453.1$ kJ on CIFAR-10 at $80\%$ and $2396.5$ kJ on CIFAR-100 at $38\%$. In these two cases, it also reaches the target faster than the corresponding $q=0$ configuration without caps.

\subsection{Sensitivity to Computation-Time Variability}
\label{sub:comp_time_sensitivity}

\RAOCSchanged{Unlike the preceding experiments, this sensitivity study considers stochastic computation times that vary across rounds and whose realizations are unknown before training. We therefore adopt a computation-first scheduling protocol rather than the chronology of Algorithm~\ref{alg:raocs}.} At each round, clients first attempt local training. The server makes the sampling decision at $T_{\mathrm{dec}}^t = \min\left\{c,\,\max_k T_k^{\mathrm{comp},t}\right\}$, where $c$ is a computation cutoff. Only clients satisfying $T_k^{\mathrm{comp},t}\leq T_{\mathrm{dec}}^t$ are eligible for sampling. \RAOCSchanged{Their freshly computed update norms are used to determine the sampling probabilities, without the $q$-refreshing mechanism.} The remaining time, $\tau-T_{\mathrm{dec}}^t$, is then used for uplink transmission.

We use CIFAR-10 with ResNet-8, $B=5$ resource blocks, round deadline $\tau=6$ s, and computation cutoff $c=4.5$ s. \RAOCSchanged{Here, $\tau$ denotes the total computation and uplink deadline, rather than the effective uplink window used in the preceding experiments.} We consider clients having the same mean computation time $\mu_{\mathrm{comp}}=3.5$ s, while the standard deviation is varied in $\sigma_{\mathrm{comp}}\in\{0.25,0.5,1.0,2.0,4.0\}$ s. All other settings follow the setup described earlier.

\begin{table}[h]
\centering
\caption{Sensitivity to computation-time variability on CIFAR-10 with ResNet-8.}
\label{tab:computation_time_sensitivity}
\resizebox{\columnwidth}{!}{%
\begin{tabular}{lcccc}
\toprule
$\sigma_{\mathrm{comp}}$
& Final Acc. (\%)
& Upload Window (s)
& Available Clients
& Selected Clients \\
\midrule
$0.25$ & $80.31$ & $1.94 \pm 0.12$ & $50.00 \pm 0.04$ & $5.42 \pm 2.09$ \\
$0.50$ & $80.59$ & $1.54 \pm 0.08$ & $48.85 \pm 1.08$ & $4.70 \pm 1.98$ \\
$1.00$ & $80.67$ & $1.50 \pm 0.00$ & $42.07 \pm 2.51$ & $4.68 \pm 1.93$ \\
$2.00$ & $80.22$ & $1.50 \pm 0.00$ & $34.67 \pm 3.18$ & $4.61 \pm 1.88$ \\
$4.00$ & $79.59$ & $1.50 \pm 0.00$ & $29.96 \pm 3.25$ & $4.57 \pm 1.88$ \\
\bottomrule
\end{tabular}%
}
\end{table}

Table~\ref{tab:computation_time_sensitivity} shows that increasing $\sigma_{\mathrm{comp}}$ mainly reduces the available client pool, from almost all clients at $\sigma_{\mathrm{comp}}=0.25$ s to about $30$ clients at $\sigma_{\mathrm{comp}}=4.0$ s. However, the number of selected clients remains relatively stable, around $4.5$-$5.4$, and the final accuracy stays above $80\%$ except under the highest variability. This indicates that RAOCS is robust to moderate variability in computation time, while very large variability can reduce the candidate pool sufficiently to slightly degrade performance.

\section{Conclusion}

This paper proposed RAOCS, a resource-aware optimal client sampling framework for wireless federated learning. Unlike standard OCS, which minimizes sampling variance under a client-count budget, RAOCS minimizes partial-participation variance under uplink resource-block, upload-deadline, update-information, and energy constraints. Each client's wireless cost is modeled through its computation time, channel-dependent uplink rate, and minimum resource-block requirement for deadline-feasible upload. This leads to a capped water-filling rule that favors statistically important clients while accounting for wireless cost and energy feasibility. Extensive experimental results validated the theoretical analysis and demonstrated the effectiveness and efficiency of RAOCS. The optimization design is orthogonal to most existing resource allocation techniques for FL systems (e.g., transmission power) and can be combined with them to further reduce cost.

\bibliographystyle{IEEEtran}
\bibliography{references}

@inproceedings{wang2023delta,
  title     = {DELTA: Diverse Client Sampling for Faster Federated Learning},
  author    = {Wang, Lin and Guo, Yongxin and Lin, Tao and Tang, Xiaoying},
  booktitle = {Advances in Neural Information Processing Systems},
  volume    = {36},
  year      = {2023}
}

@inproceedings{mcmahan2017communication,
  title={Communication-efficient learning of deep networks from decentralized data},
  author={McMahan, Brendan and Moore, Eider and Ramage, Daniel and Hampson, Seth and y Arcas, Blaise Aguera},
  booktitle={Artificial intelligence and statistics},
  pages={1273--1282},
  year={2017},
  organization={Pmlr}
}

@article{kairouz2021advances,
  title={Advances and open problems in federated learning},
  author={Kairouz, Peter and McMahan, H Brendan},
  journal={Foundations and trends in machine learning},
  volume={14},
  number={1-2},
  pages={1--210},
  year={2021},
  publisher={Emerald Publishing Limited}
}

@article{bonawitz2019towards,
  title={Towards federated learning at scale: System design},
  author={Bonawitz, Keith and Eichner, Hubert and Grieskamp, Wolfgang and Huba, Dzmitry and Ingerman, Alex and Ivanov, Vladimir and Kiddon, Chloe and Kone{\v{c}}n{\`y}, Jakub and Mazzocchi, Stefano and McMahan, Brendan and others},
  journal={Proceedings of machine learning and systems},
  volume={1},
  pages={374--388},
  year={2019}
}

@inproceedings{rizk2021optimal,
  title={Optimal importance sampling for federated learning},
  author={Rizk, Elsa and Vlaski, Stefan and Sayed, Ali H},
  booktitle={ICASSP 2021-2021 IEEE International Conference on Acoustics, Speech and Signal Processing (ICASSP)},
  pages={3095--3099},
  year={2021},
  organization={IEEE}
}

@article{rizk2022federated,
  title={Federated learning under importance sampling},
  author={Rizk, Elsa and Vlaski, Stefan and Sayed, Ali H},
  journal={IEEE Transactions on Signal Processing},
  volume={70},
  pages={5381--5396},
  year={2022},
  publisher={IEEE}
}

@article{chen2022optimal,
  title   = {Optimal Client Sampling for Federated Learning},
  author  = {Chen, Wenlin and Horv{\'a}th, Samuel and Richt{\'a}rik, Peter},
  journal = {Transactions on Machine Learning Research},
  year    = {2022},
  month   = aug,
  issn    = {2835-8856},
  url     = {https://openreview.net/forum?id=8GvRCWKHIL}
}

@article{ribero2022f3ast,
  title={Federated learning under intermittent client availability and time-varying communication constraints},
  author={Ribero, M{\'o}nica and Vikalo, Haris and De Veciana, Gustavo},
  journal={IEEE Journal of Selected Topics in Signal Processing},
  volume={17},
  number={1},
  pages={98--111},
  year={2022},
  doi={10.1109/JSTSP.2022.3224590},
}

@inproceedings{zhang2024optimalvariance,
  title={Optimal Variance-Reduced Client Sampling for Multiple Models Federated Learning},
  author={Zhang, Haoran and Li, Zekai and Gong, Zejun and Siew, Marie and Joe-Wong, Carlee and El-Azouzi, Rachid},
  booktitle={2024 IEEE 44th International Conference on Distributed Computing Systems (ICDCS)},
  pages={1446--1447},
  year={2024},
  doi={10.1109/ICDCS60910.2024.00144}
}

@misc{zhang2025towards,
  title={Towards Optimal Heterogeneous Client Sampling in Multi-Model Federated Learning},
  author={Zhang, Haoran and Gong, Zejun and Li, Zekai and Siew, Marie and Joe-Wong, Carlee and El-Azouzi, Rachid},
  year={2025},
  eprint={2504.05138},
  archivePrefix={arXiv},
  primaryClass={cs.LG}
}

@article{zhao2025adaptive,
  title={Adaptive client sampling in federated learning via online learning with bandit feedback},
  author={Zhao, Boxin and Wang, Lingxiao and Liu, Ziqi and Zhang, Zhiqiang and Zhou, Jun and Chen, Chaochao and Kolar, Mladen},
  journal={Journal of Machine Learning Research},
  volume={26},
  number={8},
  pages={1--67},
  year={2025}
}

@inproceedings{nishio2019client,
  title={Client selection for federated learning with heterogeneous resources in mobile edge},
  author={Nishio, Takayuki and Yonetani, Ryo},
  booktitle={IEEE international conference on communications (ICC 2019)},
  pages={1--7},
  year={2019},
  doi={10.1109/ICC.2019.8761315}
}

@article{huang2020efficiency,
  title={An efficiency-boosting client selection scheme for federated learning with fairness guarantee},
  author={Huang, Tiansheng and Lin, Weiwei and Wu, Wentai and He, Ligang and Li, Keqin and Zomaya, Albert Y},
  journal={IEEE Transactions on Parallel and Distributed Systems},
  volume={32},
  number={7},
  pages={1552--1564},
  year={2020},
}

@article{xu2020client,
  title={Client selection and bandwidth allocation in wireless federated learning networks: A long-term perspective},
  author={Xu, Jie and Wang, Heqiang},
  journal={IEEE Transactions on Wireless Communications},
  volume={20},
  number={2},
  pages={1188--1200},
  year={2020},
  doi={10.1109/TWC.2020.3031503}
}

@inproceedings{luo2022tackling,
  title={Tackling system and statistical heterogeneity for federated learning with adaptive client sampling},
  author={Luo, Bing and Xiao, Wenli and Wang, Shiqiang and Huang, Jianwei and Tassiulas, Leandros},
  booktitle={IEEE INFOCOM 2022 - IEEE conference on computer communications},
  pages={1739--1748},
  year={2022},
  doi={10.1109/INFOCOM48880.2022.9796935}
}

@article{luo2024adaptive,
  title={Adaptive heterogeneous client sampling for federated learning over wireless networks},
  author={Luo, Bing and Xiao, Wenli and Wang, Shiqiang and Huang, Jianwei and Tassiulas, Leandros},
  journal={IEEE Transactions on Mobile Computing},
  volume={23},
  number={10},
  pages={9663--9677},
  year={2024},
  doi={10.1109/TMC.2024.3368473}
}

@article{chen2021convergence,
  title={Convergence Time Optimization for Federated Learning over Wireless Networks},
  author={Chen, Mingzhe and Poor, H. Vincent and Saad, Walid and Cui, Shuguang},
  journal={IEEE Transactions on Wireless Communications},
  volume={20},
  number={4},
  pages={2457--2471},
  year={2021},
  doi={10.1109/TWC.2020.3042530},
}

@misc{vardhan2025client,
  title={Client selection in federated learning with data heterogeneity and network latencies},
  author={Vardhan, Harsh and Yu, Xiaofan and Rosing, Tajana and Mazumdar, Arya},
  year={2025},
  eprint={2504.01921},
  archivePrefix={arXiv},
  primaryClass={cs.LG},
}

@misc{ribero2020communication,
  title        = {Communication-Efficient Federated Learning via Optimal Client Sampling},
  author       = {Ribero, Monica and Vikalo, Haris},
  year         = {2020},
  eprint       = {2007.15197},
  archivePrefix = {arXiv},
  primaryClass = {cs.LG},
}

@inproceedings{hosseinzadeh2022communication,
  title={Communication-loss trade-off in federated learning: A distributed client selection algorithm},
  author={Hosseinzadeh, Minoo and Hudson, Nathaniel and Heshmati, Sam and Khamfroush, Hana},
  booktitle={2022 IEEE 19th annual consumer communications \& networking conference (CCNC)},
  pages={1--6},
  year={2022},
  doi={10.1109/CCNC49033.2022.9700601}
}

@article{li2025optimal,
  title={Optimal client selection of federated learning based on compressed sensing},
  author={Li, Qing and Lyu, Shanxiang and Wen, Jinming},
  journal={IEEE Transactions on Information Forensics and Security},
  volume={20},
  pages={1679--1694},
  year={2025},
  doi={10.1109/TIFS.2025.3526050}
}

\appendices

\section{Derivation of the Sampling Variance}
\label{app:variance_derivation}

Using the indicator $\mathbf{1}_{k\in\mathcal{A}^t}$, the sampled update is written as
\begin{equation}
    \widehat{U}^t
    =
    \sum_{k=1}^{K}
    \mathbf{1}_{k\in\mathcal{A}^t}
    \frac{\alpha_k}{p_k^t}
    U_k^t.
\end{equation}
Using the unbiasedness result in \eqref{eq:fullparticipation}, the sampling variance is
\begin{equation}
    \mathcal{V}^t(p^t)
    =
    \mathbb{E}_{\mathcal{A}^t}
    \left[
    \left\|
    \widehat{U}^t
    -
    U_{\mathrm{full}}^t
    \right\|^2
    \right].
\end{equation}
Under independent Bernoulli sampling, the indicators
$\mathbf{1}_{k\in\mathcal{A}^t}$ are independent and the cross terms vanish. Thus,
{
\footnotesize
\begin{equation}
    \mathcal{V}^t(p^t) =
    \sum_{k=1}^{K} \mathrm{Var} \left( \mathbf{1}_{k\in\mathcal{A}^t} \frac{\alpha_k}{p_k^t}
    U_k^t \right) =
    \sum_{k=1}^{K} \frac{\alpha_k^2}{(p_k^t)^2} \mathrm{Var} \left(\mathbf{1}_{k\in\mathcal{A}^t} \right) \|U_k^t\|^2.
\end{equation}
}
Since $\mathrm{Var} \left(\mathbf{1}_{k\in\mathcal{A}^t}\right) = p_k^t(1-p_k^t)$, we obtain
\begin{equation}
    \mathcal{V}^t(p^t)
    =
    \sum_{k=1}^{K}
    \alpha_k^2
    \frac{1-p_k^t}{p_k^t}
    \|U_k^t\|^2.
\end{equation}

\section{KKT Derivation of the Capped Water-Filling Solution}
\label{app:kkt_solution}

We derive the solution of \eqref{eq:main_problem}. For compactness, we omit the round index $t$. The problem is
\begin{align}
    \min_{p} \quad &
    \sum_{k=1}^{K}
    \alpha_k^2
    \frac{1-p_k}{p_k}
    \|U_k\|^2 \\
    \text{s.t.} \quad &
    \sum_{k=1}^{K} b_{k,\min}p_k \leq B, \\
    & 0 < p_k \leq \bar{p}_k,
    \quad \forall k\in\mathcal{K}.
\end{align}

For finite $b_{k,\min}$ and $p_k>0$, the Lagrangian is
{\footnotesize
\begin{equation}
    \mathcal{L}
    =
    \sum_{k=1}^{K}
    \alpha_k^2
    \left(\frac{1-p_k}{p_k}\right)
    \|U_k\|^2
    +
    \lambda
    \left(
    \sum_{k=1}^{K} b_{k,\min}p_k - B
    \right)
    +
    \sum_{k=1}^{K}\eta_k(p_k-\bar{p}_k),
\end{equation}
}
where $\lambda\geq0$ is associated with the resource-block constraint and $\eta_k\geq0$ is associated with the upper-bound constraint $p_k\leq\bar{p}_k$.

The stationarity condition is
\begin{equation}
    -\frac{\alpha_k^2\|U_k\|^2}{p_k^2}
    +
    \lambda b_{k,\min}
    +
    \eta_k
    =
    0.
\end{equation}

For a non-saturated client, $p_k<\bar{p}_k$, complementary slackness gives $\eta_k=0$. Hence,
\begin{equation}
    p_k
    =
    \frac{\alpha_k\|U_k\|}
    {\sqrt{\lambda b_{k,\min}}}.
\end{equation}
For a saturated client, the upper bound is active and $p_k=\bar{p}_k$.

Let $\mathcal{O}$ denote the set of non-saturated clients. Then
\begin{equation}
    p_k^\star
    =
    \begin{cases}
    \displaystyle
    \frac{\alpha_k\|U_k\|}
    {\sqrt{\lambda b_{k,\min}}},
    & k\in\mathcal{O}, \\[2ex]
    \bar{p}_k,
    & k\notin\mathcal{O}.
    \end{cases}
\end{equation}

When the resource constraint is active, we have
\begin{equation}
    \sum_{k=1}^{K} b_{k,\min}p_k^\star = B.
\end{equation}
Substituting the piecewise form gives
\begin{equation}
    \sum_{k\in\mathcal{O}}
    b_{k,\min}
    \frac{\alpha_k\|U_k\|}
    {\sqrt{\lambda b_{k,\min}}}
    +
    \sum_{k\notin\mathcal{O}}
    \bar{p}_k b_{k,\min}
    =
    B.
\end{equation}
Thus,
\begin{equation}
    \frac{1}{\sqrt{\lambda}}
    =
    \frac{
    B-\sum_{j\notin\mathcal{O}} \bar{p}_j b_{j,\min}
    }{
    \sum_{j\in\mathcal{O}}
    \alpha_j\|U_j\|\sqrt{b_{j,\min}}
    }.
\end{equation}

Therefore, for $k\in\mathcal{O}$,
\begin{equation}
    p_k^\star
    =
    \frac{
    B-\sum_{j\notin\mathcal{O}} \bar{p}_j b_{j,\min}
    }{
    \sum_{j\in\mathcal{O}}
    \alpha_j\|U_j\|\sqrt{b_{j,\min}}
    }
    \cdot
    \frac{\alpha_k\|U_k\|}
    {\sqrt{b_{k,\min}}},
\end{equation}
while $p_k^\star=\bar{p}_k$ for $k\notin\mathcal{O}$. The  set $\mathcal{O}$ is characterized by the non-saturation condition $p_k^\star<\bar{p}_k$, as characterized in equation  (\ref{eq:ocs_resource_set}).

\end{document}